\documentclass[sigconf]{acmart}

\usepackage{booktabs} 
\usepackage{enumitem}
\usepackage{url}
\usepackage{setspace}

\usepackage{algorithm}
\usepackage{algpseudocode}
\usepackage{mathtools}

\newif\ifdraft\drafttrue

\ifdraft

\newcommand\todo[1]{{\footnotesize \color{red}[#1 - \textbf{TODO}]}}
\newcommand\tf[1]{{\footnotesize \color{green}[#1 - \textbf{Tobi}]}}
\newcommand\mz[1]{{\footnotesize \color{blue}[#1 - \textbf{Martina}]}}
\newcommand\hjc[1]{{\footnotesize \color{yellow}[#1 - \textbf{Hyung Jin}]}}
\newcommand\ac[1]{{\footnotesize \color{blue}[#1 - \textbf{Antoine}]}}
\newcommand\mpet[1]{{\footnotesize \color{magenta}[#1 - \textbf{Max}]}}
\newcommand\yd[1]{{\footnotesize \color{red}[#1 - \textbf{Yiannis}]}}
\else
\newcommand\tf[1]{}
\newcommand\mz[1]{}
\newcommand\hjc[1]{}
\newcommand\ac[1]{}
\newcommand\mpet[1]{}
\newcommand\yd[1]{}
\newcommand\todo[1]{}
\fi

 \copyrightyear{2019} 
 \acmYear{2019} 
 \setcopyright{acmlicensed}
\acmConference[GECCO '19]{Genetic and Evolutionary Computation Conference}{July 13--17, 2018}{Prague, Czech Republic}
 \acmPrice{15.00}
 \acmDOI{10.1145/3321707.3321804}
 \acmISBN{978-1-4503-6111-8/19/07}

\newcommand\algoname{AURORA}

\begin{document}
\title{Autonomous skill discovery with Quality-Diversity and Unsupervised Descriptors}

\author{Antoine Cully}
\orcid{0000-0002-3190-7073}
\affiliation{%
  \institution{Imperial College London}
}
\email{a.cully@imperial.ac.uk}
 
\begin{abstract}
Quality-Diversity optimization is a new family of optimization algorithms that, instead of searching for a single optimal solution to solving a task, searches for a large collection of solutions that all solve the task in a different way. This approach is particularly promising for learning behavioral repertoires in robotics, as such a diversity of behaviors enables robots to be more versatile and resilient. However, these algorithms require the user to manually define behavioral descriptors, which is used to determine whether two solutions are different or similar. The choice of a behavioral descriptor is crucial, as it completely changes the solution types that the algorithm derives. In this paper, we introduce a new method to automatically define this descriptor by combining Quality-Diversity algorithms with unsupervised dimensionality reduction algorithms. This approach enables robots to autonomously discover the range of their capabilities while interacting with their environment. The results from two experimental scenarios demonstrate that robot can autonomously discover a large range of possible behaviors, without any prior knowledge about their morphology and environment. Furthermore, these behaviors are deemed to be similar to handcrafted solutions that uses domain knowledge and significantly more diverse than when using existing unsupervised methods.

\end{abstract}


\maketitle

\section{Introduction}

\begin{figure}[!t]
\centering \includegraphics[width=0.75\columnwidth]{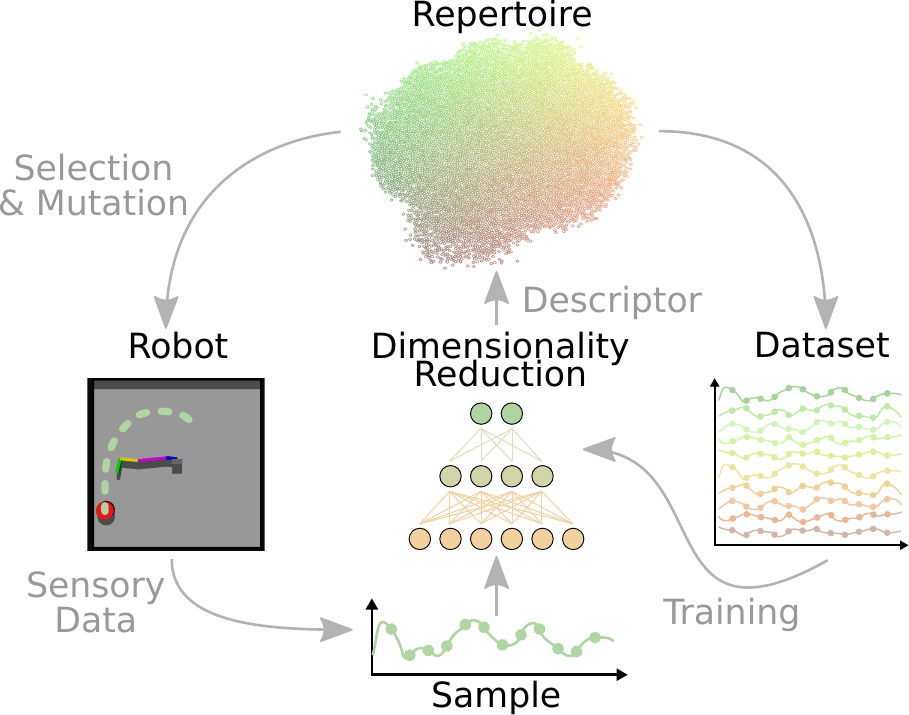}
\caption{AURORA: A quality diversity (QD) algorithm used to generate a behavioral repertoire of a robot. The behavioral descriptor used in the QD algorithm is defined by a dimensionality reduction algorithm (like an auto-encoder), which is trained and periodically refined based on the sensory data produced by the behaviors contained in the repertoire. The resulting approach enables robots to autonomously discover the range of their abilities.}
\label{fig:concept}
\end{figure}


A large field of robotics research is devoted to the design of robots that can support people in their everyday duties. However, one aspect that prevents robots from being largely adopted in our daily life is their inability to adapt to and overcome unforeseen situations. It is impossible for engineers to anticipate all the situations that a robot will encounter, particularly when they are not operating in well-structured environments, like factories. To address this issue, a new approach has been proposed, which involves using learning algorithms that enable robots to learn how to overcome new situations, like mechanical damages \cite{cully2015robots,bongard2006resilient}. However, engineers and technical experts are still required in this approach to define fitness and reward functions, or other aspects of learning algorithms. This limitation is tractable in industrial contexts, but it is unthinkable to require a roboticist to re-configure the learning system every time something changes in a domestic household.

In some settings, prior knowledge and expertise simply do not exist. Examples include the design of new robot morphologies~\cite{nygaard2018real} or soft robots with automated tools~\cite{corucci2018evolving,cheney2013unshackling}, where it is particularly challenging to know in advance the capabilities of the robots. A large amount of experiments and evaluations is required to acquire this knowledge. Similarly, it is difficult to know beforehand how a robot would behave in a completely new environment. For example, what would the behavior of a legged robot be when walking on snow, or on ice~\cite{corucci2018evolving}? The main issue is that a high level of expertise and prior knowledge is usually required to configure and fine-tune learning algorithms, so as to avoid undesired behaviors or guarantee high-performance (for instance, with helper objectives~\cite{doncieux2014beyond}). 

To address these challenges, we present AURORA: AUtonomous RObots that Realize their Abilities.
AURORA combines Quality-Diversity (QD) algorithms and dimensionality reduction (DR) algorithms to leverage their strengths and mitigate their weaknesses. On one hand, QD algorithms have shown to be instrumental in generating large collections of solutions~\cite{cully2017quality,pugh2016quality}, often leading to surprisingly creative results~\cite{lehman2018surprising, cully2015robots}. However, it requires the definition of a Behavioral Descriptor (BD) to discriminate the different families of solutions and drive the exploration. The selection of the BD remains crucial and challenging, as it is particularly task dependent and alter the solutions the QD algorithm derives~\cite{pugh2015confronting}. On the other hand, DR algorithms, like deep auto-encoders or PCA, are powerful tools to extract meaningful low-dimensional representations from high-dimensional data spaces. Nevertheless, they usually require a large amount of training data, which can be challenging in certain applications.
AURORA compensates for the limitations of both approaches by exploiting each of their respective advantages: the large collection of solutions generated by the QD algorithm is used as a training dataset for the DR algorithm, and the latent representation learned by the DR serves as a BD for the QD algorithm (see Fig.~\ref{fig:concept}). 
This mutual connection enables AURORA to autonomously generate behavioral repertoires without the need to manually define the BDs or engineer insights about the task or the robot capabilities. In this sense, robots can autonomously discover the range of their capabilities through self-exploration.


The performance of AURORA is evaluated on two experimental scenarios in which robots have to learn how to interact with their environment (i.e., throwing a projectile or pushing a puck). In both scenarios, the results show that the collections of behaviors (also called Behavioral Repertoires) generated by AURORA are at least as diverse as those created by manually defined BDs (including the help of prior knowledge about the robot and its environment) and significantly better than existing unsupervised approaches. In addition to the benefits of this approach on the autonomy of robots, it is also a valuable tool to automatically set the BD, which is often one of the most challenging aspects to configure in QD algorithms~\cite{cully2017quality}.

\section{Related Work}
\subsection{Quality-Diversity and Behavioral Repertoires}
QD algorithms are often used in robotics to generate collections of skills that are called behavioral repertoires. The objective in this case is to create a large collection of diverse and effective controllers. Each controller, defined as a neural network or a set of parametric functions~\cite{gomes2018approach,duarte2018evolution,cully2015evolving}, governs the robot's motors to execute a behavior and potentially solve a task. Each of the produced behaviors is then represented by a numerical vector called a \emph{Behavioral Descriptor} (BD). In the vast majority of the literature, the function that assigns the BD to each observed behavior is manually and carefully selected depending on the considered tasks~\cite{cully2017quality, pugh2015confronting} (exceptions are reviewed in next sections). The choice of descriptor is crucial, as it is used to quantify the novelty of each new solution and to drive the exploration of new families of solutions~\cite{lehman2011abandoning}. 

As discussed in \cite{cully2017quality}, most of these variants can be described within a common framework composed of a \emph{container} and \emph{selector}. 
MAP-Elites~\cite{mouret2015illuminating} and Novelty Search with Local Competition (NSLC)~\cite{lehman2011evolving} are two well-known examples of QD algorithms. However, a large number of variants, as well as new applications, have been proposed over the last five years~\cite{gaier2018data,khalifa2018talakat,urquhart2018optimisation}. 
The container is used to store and organize the collection of controllers derived by the algorithm. In MAP-Elites, the container is a grid formed by the discretization of the BD-space, while in NSLC the repertoire is an unstructured archive constructed based on the distance between BDs.

QD algorithms are typically initialized by randomly generating a fixed number of controllers (e.g., 100) and storing them in the container after evaluation. During the evaluation, in addition to recording the performance of the controller, a BD is computed from sensory data (e.g., the final position of a robot after walking for three seconds, like in \cite{cully2013behavioral}). The BD is then used to sort the different behaviors in the container. 
After the initialization, the usual iteration of a QD algorithm consists of 4 steps: 1) selecting a controller from the container via the selector; 2) creating a mutated copy of this controller; 3) evaluating this new solution; and 4) attempting to store it in the container. If a controller produces a BD that already exists in the container (or a very similar one), then only the best (according to a fitness function) of the two controllers is retained. This selection mechanism improves the quality of the controllers contained in the container. In the case when there is no similar behavior already found within the container, the new controller is added to the container, consequently increasing the diversity of the container's solutions. The selection of the controller in step 1 is governed by a selector that can either select controllers according to a uniform distribution, or a biased distribution that favors certain types of solutions. For instance, the curiosity score \cite{cully2017quality} can be used to focus on solution types that produce new controllers, which are often added to the container. This selection approach has been shown to improve the coverage of the produced collections in several robotic setups~\cite{cully2017quality}. 

In this work, we consider a QD algorithm composed of an unstructured archive (like the one used in NSLC) and a selector based on the curiosity score.

\subsection{Automatic behavioral characterization}

Defining the BD has always been a challenging endeavour in the design of QD algorithms~\cite{pugh2015confronting}. It is therefore natural to see several papers propose approaches to make its definition automatic.

Meyerson et al.~\cite{meyerson2016learning} present a method that enables the automatic determination of BD by learning which descriptor usually leads to controllers that effectively solve the task. They first use some existing BDs with a default QD algorithm (novelty search) to train a population of controllers on different ``training'' tasks. During the evolutionary process, they record the behavior (i.e., some sensory data) and the fitness of each evaluated controller. Then, an algorithm uses the recorded data and learns a new BD that will lead to more successful controllers on new ``test'' tasks (not seen during the training). 
In their experiments, the resulting BDs significantly outperform manually defined ones during maze navigation tasks. While this approach is promising and leads to positive results, it requires a way of defining ``successful individuals'' to learn the corresponding BDs, which usually involves some prior knowledge about the robots or the tasks. In this work, the motivation is to enable a robot to discover its own capabilities and skills without the need of engineers or technical experts. It is therefore challenging to define success in such a scenario and apply it to Meyerson's approach without prior knowledge on the task or environment.  

Using dimensionality reduction (DR) algorithms to automatically define BDs from the robot's perceptions (i.e., sensory data) has also been investigated in recent works. For instance, an auto-encoder trained on the MNIST dataset (dataset of hand-written digits~\cite{lecun1998gradient}) has been used to enable a robotic arm to autonomously learn how to write digits from 0 to 9, while learning a hierarchical behavioral repertoire~\cite{ cully2018hierarchical}. 
In a similar context, P\'er\'e et al.~\cite{pere2018unsupervised} used intrinsically motivated goal exploration processes (a divergent search algorithm that shares similarities with QD optimization) coupled with DR algorithms to generate families of robotic controllers. Similarly to~\cite{cully2018hierarchical}, they pre-trained the DR algorithms with a dataset that has been specifically generated for the considered task. The task presented in~\cite{pere2018unsupervised} is similar to the air hockey task used in this paper, in which a robot can interact with an object (a puck, a ball, or more complicated objects) that can bounce on several walls. The dataset used to pre-train the DR algorithms has been generated to represent all the different configurations of the environment (i.e., all the possible positions of the puck). The learning algorithm is then used to discover controllers that enable the robot to push the object towards the locations defined by the dataset. 

AURORA follows the same idea of using DR algorithms to automatically define the BDs. However, instead of providing a manually generated dataset, which requires a certain amount of knowledge about the task and the different ways a robot can solve the task, we let our robots construct their own datasets, purely by interacting with their environment (like in ~\cite{liapis2013transforming}). The resulting approach is thus fully unsupervised and enables robots to explore the range of possible behaviors on their own.

\subsection{Behavioral characterization in high-dimensional spaces}\label{sec:CVT}
Instead of attempting to find a low-dimensional representation of the behaviors (which is the usual approach in the literature~\cite{lehman2011evolving, mouret2015illuminating, cully2018quality}), Vassiliades et al.~\cite{vassiliades2018using} propose to directly define the BDs in high-dimensional space. For this, they automatically generate a uniform discretization of the plausible parts of the high-dimensional sensory data space. 
With this approach, it is possible to directly consider the entire trajectory of a mobile robot, or complete time series of sensory data. This is achieved by constructing a Centroidal Voronoi Tessellation (CVT) to obtain a set of ``centroids'' that are uniformly distributed over the space of possible high-dimensional BDs. The set of centroids is then used to create a discretization of the high-dimensional behavioral space, which can be used as a grid for MAP-Elites. The construction of the CVT is usually done by uniformly sampling the space of possible high-dimensional BDs and then running a clustering algorithm (like k-means~\cite{macqueen1967some}) to find $k$ clusters that will serve as centroids. The $k$ value is a parameter defined by the user to control the number of cells (or niches) for the MAP-Elites grid. 

This approach has yielded positive results across several experimental conditions, including those with the BD-space ranging from 2 to 1000 dimensions~\cite{vassiliades2018using}. It has been used to tackle maze navigation tasks with a simulated wheeled robot, and locomotion tasks (including damage adaptation) on a simulated hexapod robot. The obtained results were significantly superior to directly using MAP-Elites for similar dimensions.  
The main reason for this performance difference is that while MAP-Elites uniformly discretizes the BD space (here corresponding to the sensory data space), CVT-MAP-Elites searches for a uniform distribution of the possible sensory-data that the robots can generate. 
For instance, if the BD is built from the robot's trajectory, which is defined as a vector of 100 dimensions (50 time-steps of 2D positions), then the BD-space is 100-dimensional. However, a randomly generated 100D vector is very unlikely to be a realistic trajectory that matches the physical constraints of the robot (e.g., the robot might be teleporting itself from one part of the space to another). Therefore, the vast majority of this space (even if appropriately bounded) will refer to trajectories that are physically impossible and many of the cells of the container would be impossible to fill. CVT-MAP-Elites, on the other hand, only samples physically plausible trajectories to construct its centroids. Therefore, all the cells used by CVT-MAP-Elites could potentially be filled. 
For instance, \cite{vassiliades2018using} reports that in a 20D maze navigation scenario, MAP-Elites manages to fill only $0.5\%$ of its cells, while CVT-MAP-Elites covers $40\%$ of them. 

While CVT-MAP-Elites is an elegant solution to scale up MAP-Elites, it makes the strong assumption that it is possible to sample the space of possible sensory data (e.g., realistic trajectories). This assumption leads to two main issues. First, it might be challenging or even impossible to sample this space without a certain level of prior knowledge regarding the task, the robot and the environment. Second, it can constrain the range of solutions that the algorithm could find. For instance, if we sample only trajectories that correspond to the robot moving forward, the algorithm will be unable to discover behaviors that make the robot move backward, even if the robot is intrinsically capable of doing it.

\section{Methods}

\subsection{\algoname}\label{sec:aurora}

AURORA works as follows: after the random initialization of the QD algorithm, the sensory data (e.g., trajectories, but any data related to the behavior can be considered) of the randomly generated controllers are collected to form a first dataset. The DR algorithm is then trained on this first dataset to learn a latent and low dimensional representation of the sensory data. The sensory data of each controller is then projected into the latent space, and the corresponding low-dimensional (e.g., 2D) projection is used as the BD. This forms the initialization procedure of AURORA. 

Then, a standard QD iteration takes place: 1) a controller is randomly selected in the collection; 2) it is mutated; 3) the resulting new controller is evaluated; and 4) attempted to be placed in the collection. The only difference from standard QD is that the BD is now automatically determined by feeding sensory data generated by this new controller into the DR algorithm and then determining the corresponding latent representation. While executing this QD iteration several times (e.g., 10000 iterations), the algorithm progressively finds more controllers to fill the latent space created by the DR algorithm. During these iterations, the number of controllers in the collection progressively increases, increasing at the same time the amount of sensory data available. After a certain number of iterations, the DR algorithm is re-trained with a new dataset composed of sensory data from all the controllers contained in the container. 

The next phase is started by updating the latent representation to take into account the new behaviors that the robot has discovered and the new sensory data that has been generated. The QD algorithm can then continue to explore the space of possible solutions with the new latent representation, which very likely offers new regions that are not already fully covered by the collection of controllers. When enough new controllers have been found, an update of the DR algorithm is performed to further refine the latent representation. The algorithm alternates between these two phases: 1) exploration via the QD iteration; and 2) latent space refinement via an update of the DR algorithm. 

After the DR update, all the controllers contained in the behavioral repertoire are assigned a new BD and stored again in the container. It often happens that some controllers with initially different BDs will be assigned updated BDs that are closer or even similar. In this case, the competition mechanism of the QD algorithm is used, and only the best ones are kept. The size of the repertoire is thus often reduced following a DR update, but quickly grows back to a large collection of solutions by finding new types of behaviors (see next section). The DR update can also change the scale of the latent space and cause an artificial growth of the repertoire size. To avoid this, the length parameter "$l$" is redefined after each update to maintain the repertoire resolution (i.e., the maximum number of controllers) constant (see ~\cite{cully2017quality} for more details). 

By updating the DR algorithm, the robot progressively acquires a better understanding of its own actions (and the corresponding observations). However, the need to update the DR algorithm varies over time. For instance, at the beginning of the learning process, the robot is very likely to observe a lot of potentially new sensory data points. It is therefore important to frequently update the latent representation and take this data into account at the beginning of the learning process. On the other hand, after an extensive period of exploration, the robot starts to have a mature internal representation of its sensory data, and is more focused on the refinement of its behavioral repertoire (i.e., increasing the quality) than on the exploration and discovery of new perceptions. At this stage of development, updating the latent space can afford to be less frequent. Following this idea, we have defined the update frequency so that it decreases exponentially over time. In the following experiments, updates take place after the following batch counts (or generation): 0, 50, 150, 350, 750, 1550, 3150 (for a total of 5000 batches).

\subsection{Dimensionality reduction algorithms}\label{sec:DR}

In the experiments presented in this paper, we consider two DR algorithms: 1) PCA; and 2) a deep auto-encoder (AE). The following two paragraphs present each method. However, other types of DR algorithms can be used with AURORA, as the only constraint is to learn a ``projector'' that can project original data onto the latent space. 

\paragraph{Principal component analysis}
PCA is a traditional approach in machine learning to find a linear projection into an orthogonal base with linearly uncorrelated variables. However, it can also be used for DR by considering only some of the most important principal components. In this work, we use this approach to project the sensory data into a low dimensional representation (i.e., 2D), which serves as a BD. 

\begin{figure}[!t]
\centering \includegraphics[width=1\columnwidth]{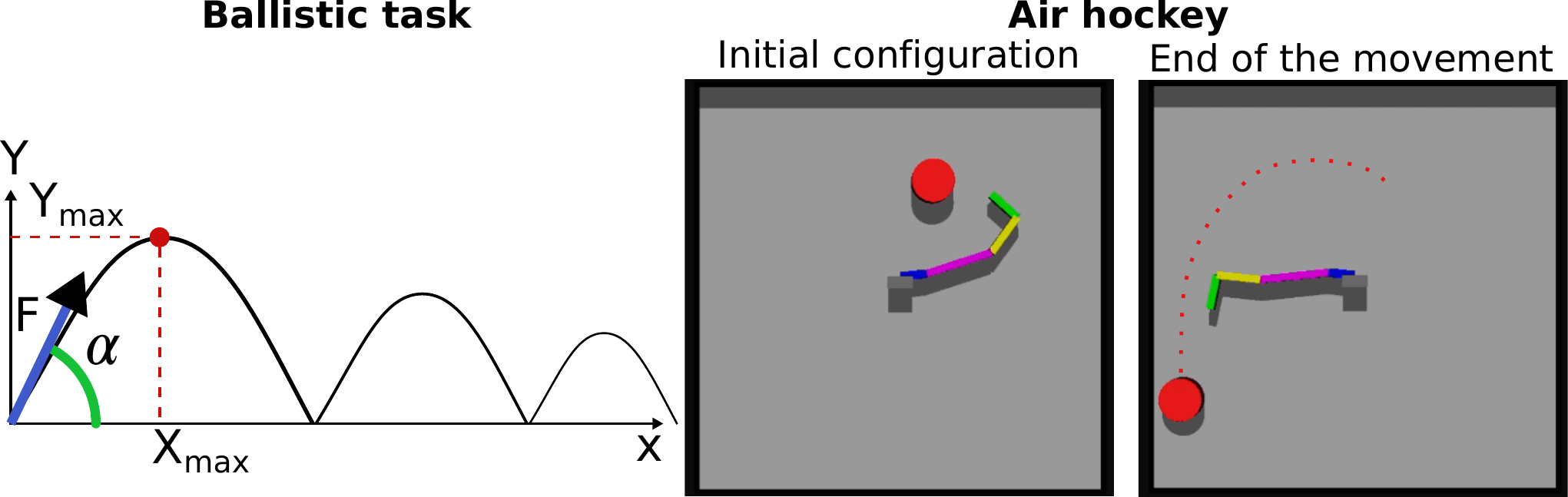}
\caption{Experimental scenarios. Ballistic task: a robot controls the magnitude (F) and the direction ($\alpha$) of the force initially applied to a projectile. Air hockey task: a four degrees of freedom robot can push a puck, which can bounce on the four walls of the arena. The goal of these experiments is to discover all the possible trajectories of the projectile or puck. }
\label{fig:setups}
\end{figure}

\paragraph{Auto-Encoders}
An AE is composed of two networks: an encoder and a decoder. The encoder takes as input some data and outputs a low-dimensional representation (i.e., a position in the latent space), while the decoder takes as input the low-dimensional representation and attempt to reconstruct the original data. The two networks are jointly trained to minimize the ``reconstruction error'', which is an absolute difference between the original data fed to the encoder and the reconstruction provided by the decoder. 
Compared to PCA, the main advantage of AE is that the projection in the latent space (as well as the reconstruction) is non-linear. This enables the DR algorithm to account for more complicated data or to have more accurate reconstruction. 
In AURORA, we use the latent position provided by the encoder as a BD for the controller (similar to~\cite{ cully2018hierarchical}). 

To avoid over-fitting on very small datasets (particularly at the beginning of the process), we split the dataset into training and validation datasets (75\%/25\%). We stop the training of the AE as soon as the validation error (averaged over the last 500 epochs) increases. We repeat this process 5 times, with different training/validation splits (in a cross-validation way) for each DR update.

\section{Experimental validation}
\begin{figure*}[!t]
\centering \includegraphics[width=1\textwidth]{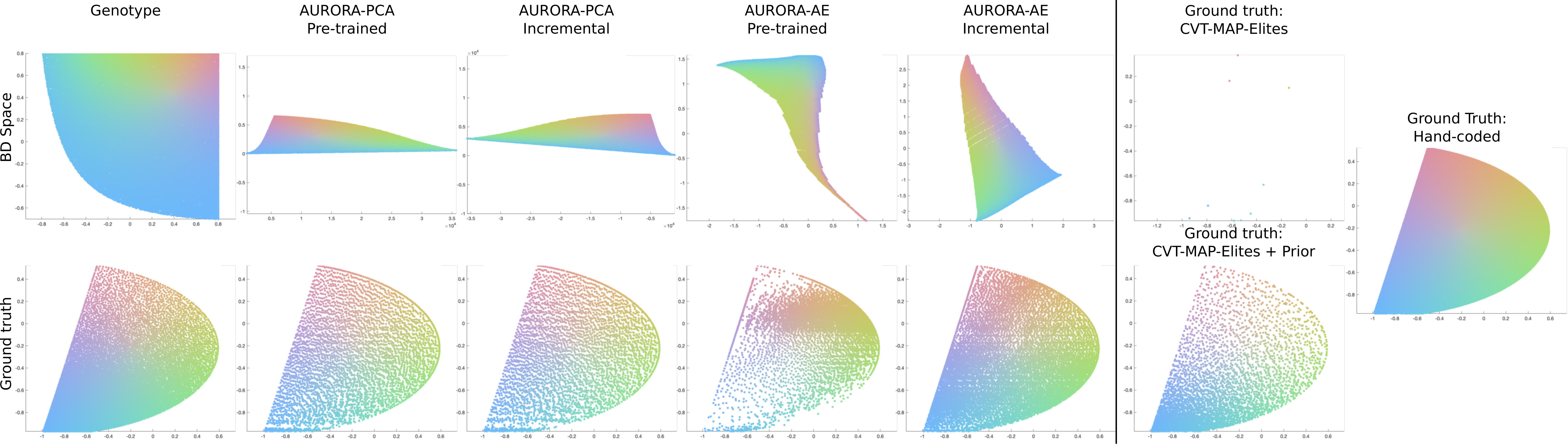}
\caption{Ballistic task. Generated BD-space compared to the ground truth. The first line represents the 2D behavioral space generated by each approach. On the second line, the behavioral repertoires are projected into the ``ground truth'' behavioral space. The ground truth behavioral space is defined by a hand-coded behavioral space. To visualize the correspondence between the different spaces, the color of each point is based on their 2D location in the ground truth BD-space.  The BD-space of CVT is not shown as it has 100 dimensions.}
\label{fig:latent_ballistic}
\end{figure*}

We evaluate the performance of AURORA in two experimental scenarios: 1) a 2D ballistic task, and 2) an air hockey task. In both tasks, we record the planar trajectories of the projectile, or puck over 50 time-steps. This forms a 100-dimensional vector, which is the sensory data fed into the DR algorithm that defines the BDs. As the main focus of AURORA is the exploration of a robot's capabilities, we do not use a fitness function in the QD algorithms (they all have a quality-score of $0$). All the approaches compared in both experimental scenarios run for 5000 batches (or generations), with 200 evaluations per batch. The QD algorithm used here is similar to the one used in \cite{cully2018hierarchical}. The experiments are replicated 20 times to generate statistical results. The AE used in all the experiments below uses the same structure: convolution layer (2 feature maps) - dense layers (5-2 neurons) - (latent space) - dense layer (5 neurons) - deconvolution layer (2 feature maps) - dense layer (100 neurons) - (output).

Given the intrinsic differences between these two tasks (e.g., the presence or absence of ground truth data), we used different types of metrics in each experiment. The following sections present the metrics, reference algorithms, and experimental results of each scenario.

\subsection{First scenario: Ballistic task}
The 2D ballistic task simulates a two degrees of freedom system that controls the direction and intensity of the force initially applied to a projectile (a punctual mass). The projectile is then released and follows a ballistic trajectory with bounces on the ground. The objective of this scenario is to quantify if AURORA is able to find all the trajectories that can be produced in this context.

This experimental setup offers several advantages. First, the computational cost to run it is particularly low, as the movement of the projectile can be expressed with simple, iterative functions, and thus does not require any computationally expensive physical simulator. Second, projectiles follow quadratic curves that can be fully described by the altitude and the distance of the highest point (see $X_{max}$ and $Y_{max}$ in Fig.~\ref{fig:setups}). We can use this as a reference BD (i.e., as a ground truth) and investigate whether it is possible to automatically generate a BD that generates a behavioral repertoire capable of covering the same range of possible trajectories.

In this experiment, the controllers are composed of two parameters, which governs the direction ($\alpha$) and the intensity (F) of the initial force applied to the projectile (see Fig~\ref{fig:setups}).

\begin{figure*}[!t]
\centering \includegraphics[width=1\textwidth]{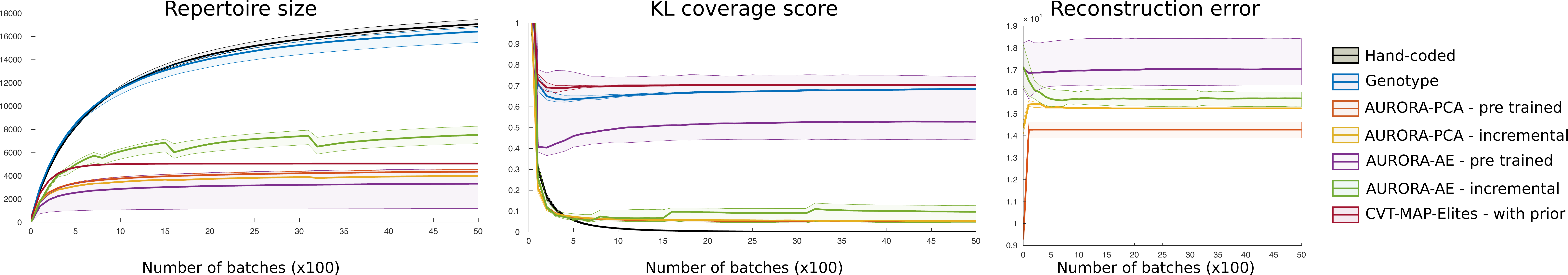}
\caption{Ballistic task. Quantitative comparison of the different algorithms based on the repertoire size, the Kullback-Leibler Coverage score, and the reconstruction error. The central line of the shaded areas represents the median value while the shaded areas extend to the first and third quartiles of the data distribution.}
\label{fig:mp_score}
\end{figure*}

\subsubsection{Performance metrics}
Quality evaluation of the produced repertoire can be done by comparing the distribution of the behaviors to the distribution obtained when using the hand-coded descriptor (i.e., $X\_max$ and $Y\_Max$ in Fig.~\ref{fig:setups}).
To compare distributions, we first project each behavior of the generated behavioral repertoires onto the ground truth (manually defined) latent space, in order to have a common representation. From this, we can use the Kullback-Leibler Coverage (KLC) introduced in \cite{pere2018unsupervised} to measure the divergence between two repertoires.
The KLC creates a normalized histogram of distributions, with 30 bins per dimension, and then computes the Kullback-Leibler divergence from each point in the histogram:
\begin{equation}
   \textrm{KLC} = \mathcal{D}_{\textrm{KL}}[E||A]=\sum_{i=1}^{30}E(i)log  \frac{E(i)}{A(i)}
\end{equation}
with $E$ representing the reference distribution and $A$ the compared distribution. In the experimental results described below, we use as a reference distribution the final BD generated from the ground-truth BD. Because this metric is based on a normalized histogram, it is not affected by a difference in the repertoire size (see Fig.~\ref{fig:mp_score} left). 

In addition to the KLC, we also report the size of the produced behavioral repertoires and the average root mean square error (RMSE) of the reconstructed trajectories from the DR algorithms. A good reconstruction ability is useful to predict consequences of the robot's actions, without the need to store all of the robot's previous observations.

\subsubsection{Compared approaches}
In this experiment, we compare eight approaches of generating the BD:
\begin{itemize}[leftmargin=*]
    \item \emph{Hand-coded} (ground truth): uses a manually defined 2D BD composed of $X_{max}$ and $Y_{max}$ described in Fig.~\ref{fig:setups}.
    \item \emph{Genotype}: uses the parameters of the controller (i.e., the genotype) as a BD. In this experiment, the genotype is 2D and can be used directly as a BD. 
    \item \emph{AURORA-PCA (pre-trained and incremental)}: uses a PCA to project the projectile trajectory onto a 2D representation (based on the two principal components). Two variants are considered. In the pre-trained variant, a dataset with 10000 samples is generated by uniformly sampling the possible actions (with 100 values per dimension). This dataset is used to pre-train the PCA, which then remains unchanged during the evolutionary process. In the ``incremental'' variant, the PCA is trained and updated during the evolutionary process. The first PCA is computed from the random controllers generated during the initialization of the QD algorithm, and is then updated periodically as described in section~\ref{sec:aurora}.
    \item \emph{AURORA-AE (pre-trained and incremental)}: uses an AE to project the trajectory onto a 2D latent space, which serves as a BD-space. Similarly to the PCA variants, the ``pre-trained'' variant,  a dataset is used to train the AE before the evolutionary process. The AE is not updated during the rest of the process. In the ``incremental'' variant, the AE is initialized and progressively refined with the controllers found by the QD algorithms, as explained in section~\ref{sec:aurora}. 
    \item \emph{CVT-MAP-Elites (with and without prior)}: uses CVT-MAP-Elites~\cite{vassiliades2018using} to directly run a QD algorithm with a high-dimensional BD-space (100 dimensions). However, as described in section~\ref{sec:CVT}, CVT-MAP-Elites requires some prior knowledge about the data distribution in order to sample feasible trajectories. In the ``with prior'' variant, the dataset used in the pre-trained variants of PCA and AE is used to initialize the CVT representation (with 10000 centroids). In the ``without prior'' variant, we do not provide this prior knowledge to the algorithm, which then samples the full space of 100 dimensions (bounded by the highest values recorded along the x and y axes of this experiment, and with 100000 centroids). This variant uses the same, minimal, amount of prior knowledge as the incremental variants of PCA and AE.

\end{itemize}

\begin{figure*}[!t]
\centering \includegraphics[width=0.85\textwidth]{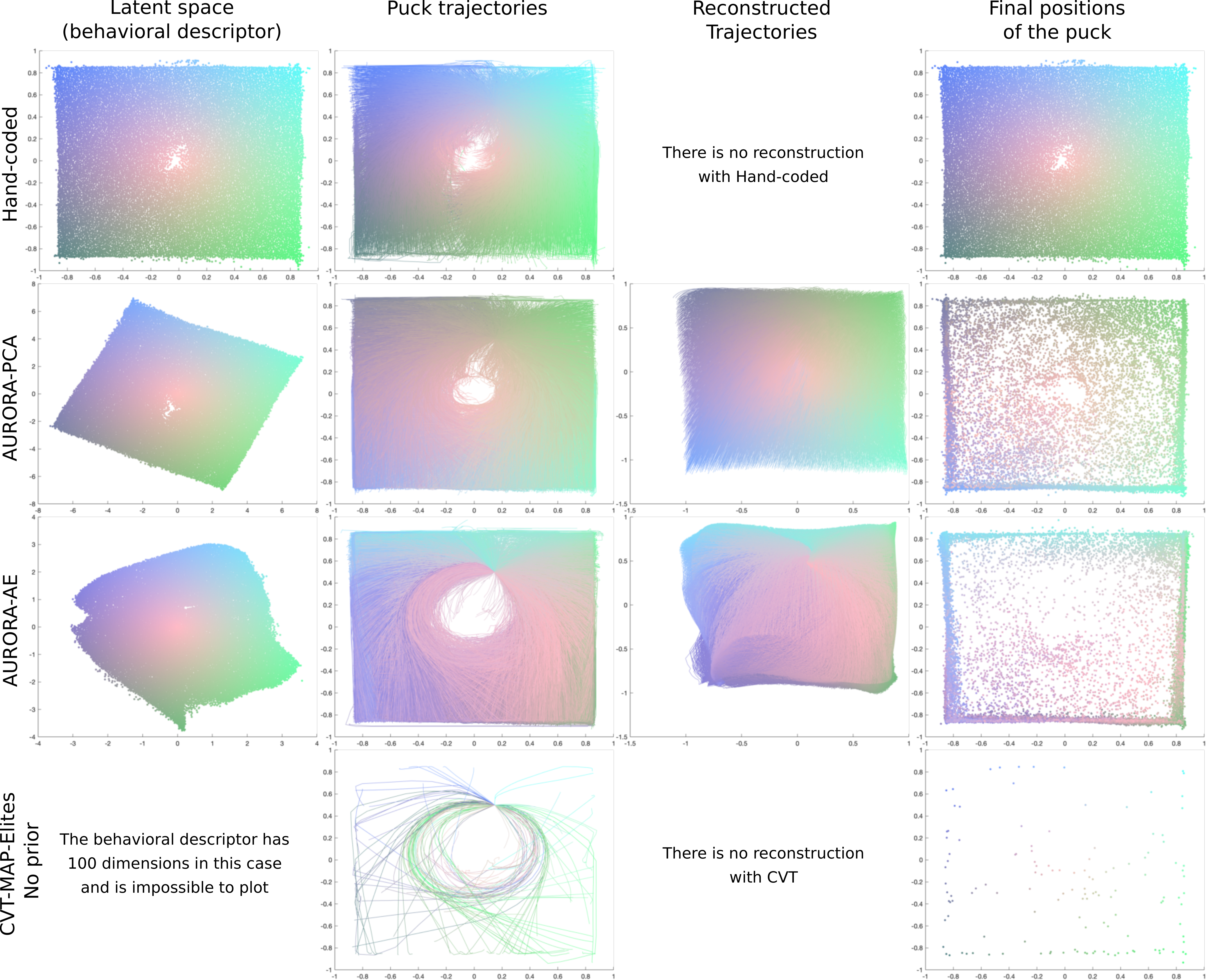}
\caption{Air hockey task. Comparison of the produced BD-spaces, real trajectories, reconstructed trajectories and final locations of the puck. To visualize the correspondence between the different spaces, the color of each point and trajectory is based on their corresponding location in the BD-space. }
\label{fig:latent_arm}
\end{figure*}

\subsubsection{Experimental results: }
Fig.~\ref{fig:latent_ballistic} shows the BD-spaces generated by each compared approach. 
We observe that all the compared approaches, excluding CVT-MAP-Elites without prior, manage to generate behaviors in all the reachable regions of the ground truth BD-space. It is expected that without a prior, CVT-MAP-Elites will perform poorly. Indeed, even when discretized with 100000 bins, the 100-dimensional space only contains a few bins that actually represent possible trajectories.

The main difference between all the other approaches is the density of each region. For instance, we can observe that the Genotype variant over-represents the blue region of the ground truth space (i.e., the blue region is denser than the rest of the ground truth latent space). This is caused by the fact that a certain part of the action space (which here corresponds to the genotype space) leads to a smaller difference in the produced trajectories than other parts of the action space. The consequence of this over-representation is that certain types of behaviors will be less represented in the behavioral repertoire (and thus offer a lower resolution of the possible diversity) than other types of solutions. On the other hand, we see that both AURORA-PCA variants and the AURORA-AE incremental manage to accurately match the distribution of the ground truth BD-space. The AURORA-AE pre-trained presents a more dense distribution in certain areas, which might be caused by the fact that for this variant, the training/validation split discussed in section~\ref{sec:DR} was not used, as the dataset was relatively large. This might have caused some over-fitting leading to an over-representation of some, probably more challenging, regions of the sensory-data space. 

This is also represented in the KLC scores of Fig~\ref{fig:mp_score}. We can see that hand-coded BD (which corresponds to the ground truth representation) eventually converges to zero, as expected. The scores of the two AURORA-PCA variants overlap with each other and cannot be distinguished in the figure, but perform well according to their final median performances of $0.051$ and $0.053$. Similarly, AURORA-AE incremental performs well with a median value of $0.097$.
The genotype, AURORA-AE pre-trained and CVT-MAP-Elites with prior present a higher KLC (respective medians: $ 0.686$, $0.528$ and $0.704$), which reflects the over/under-representation of certain regions of the BD-space.
The CVT without prior variant is not displayed in this figure, as it performs very poorly (with a median of 3.2, which is more than 30 times worse than the AURORA-PCA or AURORA-AE incremental approaches).

The reconstruction errors of the PCA variants are slightly better than the AE variants (the reconstruction error is not defined for the other approaches). This might mean that the trajectories of this task can be captured with a linear combination of features. The next experiment involves more complex trajectories and shows different results in terms of reconstruction error. 

 From this first experiment, we can observe that the quality of the repertoires produced by AURORA are close to repertoires derived from manually defined BDs, while not requiring any prior knowledge about the type of data used or any intelligent pre-sampling, like in CVT-MAP-Elites. This demonstrates that an unsupervised DR algorithm can be used to project sensory data (like trajectories) in a latent space and use it as a BD.

\subsection{Second scenario: Air hockey task}

The second experimental setup is an air hockey task. A planar robotic arm, with 4 degrees of freedom that can push a puck in multiple directions and make it bounce on the four walls of the environment (see Fig~\ref{fig:setups}).
The controllers are defined by eight angular positions, that correspond to the initial and final positions of the four robot joints. The robot first goes to its initial position, the puck is placed in the environment, and finally the robot moves to its final configuration, eventually pushing the puck. The objective here is to analyze the diversity of the trajectories that a robot will be able to autonomously generate when the descriptor is automatically defined. This task is significantly more challenging than the previous one, mainly because in this task all the possible trajectories that the robot can generate in this environment are not known (or this will require a large amount of engineering and prior knowledge). The search space is also larger and the produced trajectories are more complicated. The goal of this experiment is to study the diversity of the trajectories that the robot can discover in a completely unsupervised approach, without any prior knowledge about itself or the environment. 

\begin{figure*}[!t]
\centering \includegraphics[width=1\textwidth]{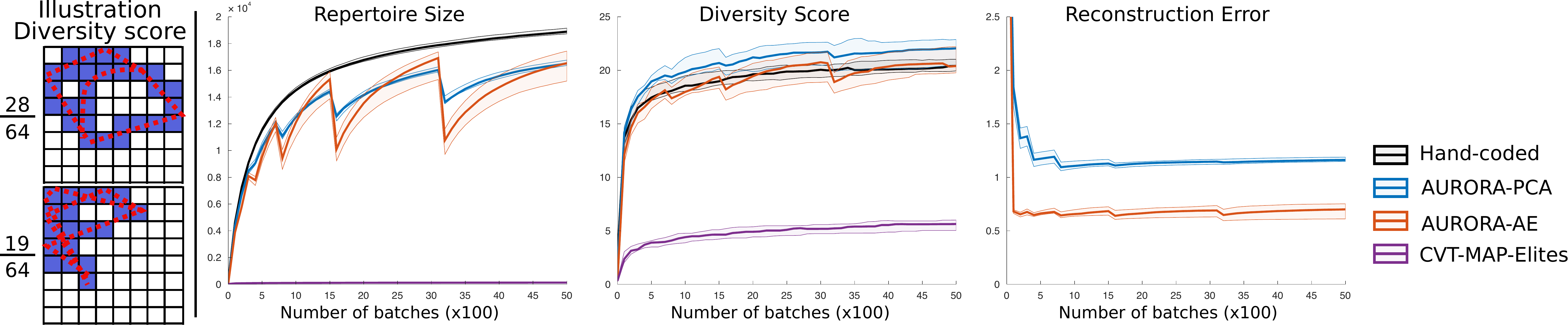}
\caption{Left, illustration of the diversity score. Right, score of the air hockey task. The repertoire size, diversity score and reconstruction error of each compared approach is represented with respect to the number of batches. This figure uses the same notations as Fig.~\ref{fig:mp_score}.}
\label{fig:arm_score}
\end{figure*}

\subsubsection{Performance metrics}
Evaluating the performance of the variant is also more challenging in this experiment as there is no reference distribution to compute the KLC.
For this experiment we therefore use another evaluation metric, which measures the diversity of the produced trajectories. The area of the robot arena (see Fig~\ref{fig:setups}) is discretized into 100 bins (10 per dimension). The trajectories from each repertoire are classified according to the bin in which the puck ends its trajectory. Then for each set of trajectories (corresponding to the different bins), we compute the number of bins that are traversed by at least one trajectory. For instance, Fig~\ref{fig:arm_score}-left shows two examples considering the same bin with three different reaching trajectories. The blue bins are traversed by at least one trajectory. The number of traversed bins divided by the total number of bins quantifies the diversity of the trajectories. The diversity score used in the following paragraphs sums this ratio over all the bins:

\begin{equation}
   \textrm{diversity} = \sum_{i=1}^N\frac{\textrm{trav\_bins}(i)}{N}
\end{equation}

\noindent with $N$ being the total number of bins (here 100), and $\textrm{trav\_bins}(i)$ the number of bins traversed by trajectories reaching the bin $i$. This score captures both the coverage of the repertoire, i.e., the number of reachable bins, and the diversity of the approaches that have been found to reach them.

\subsubsection{Compared approaches}
We consider the same variants as in the previous experiment. However, as there is no ground truth and no simple ways of yielding prior knowledge, we only evaluate the approaches that are completely agnostic about the robot and its environment: PCA incremental, Auto-encoder incremental, and CVT-MAP-Elites without prior. We also compare to a hand-coded BD that considers the final 2D location of the puck. This last variant is used as a reference to compare the different approaches with a BD that has been hand-crafted to maximize the coverage of the behavioral repertoire.

\subsubsection{Experimental results}

Fig.~\ref{fig:latent_arm} shows the learned BD spaces for each of the variants, as well as the corresponding trajectories, reconstructed trajectories and final locations of the puck for each trajectory. First, we can observe that both AURORA-PCA and AURORA-AE find a large diversity of trajectories covering the whole reachable space. We can also observe that CVT-MAP-Elites without prior is again, and for the same reasons, unable to find a lot of controllers. When comparing with a hand-coded descriptor, we can see that the density of the final puck locations are more uniform than those of AURORA-PCA or AURORA-AE. This can be explained by the fact that the hand-coded BD is explicitly defined to achieve this type of result, while AURORA finds this autonomously. We also note that AURORA presents a higher density of solutions near the borders of the arena. This might be explained as collisions with the wall absorb some kinetic energy and thus makes it easier to stop the puck around these locations. It is thus easier to generate a large number of diverse trajectories surrounding a wall than in the middle of the arena. 

Fig.~\ref{fig:arm_score}-right describes the diversity scores obtained by the different variants. We observe that CVT-MAP-Elites shows the worst score, while AURORA-AE and the hand-coded BD show similar results (the difference is not statistically significant p-value$=0.561$ with rank-sum test). AURORA-PCA performs slightly better than these two other approaches (p-value $<0.038$). 
We can also observe that the reconstruction error is lower with AURORA-AE than with AURORA-PCA. This is mainly because the trajectories are highly non-linear in this experiment, which prevents PCA from producing accurate reconstructions. 

The fact that PCA usually shows a slightly better coverage or diversity in both experiments is most likely due to the fact that PCA searches for the two main linear combinations that capture most of the variance in the data. On the other hand, the definition of the latent space in an AE is only governed by how it might help the reconstruction of the trajectories. There is no incentive in a traditional AE to structure the latent space according to a specific criterion. In future works, we will investigate the use of Variational AE~\cite{kingma2013auto} or t-SNE\cite{maaten2008visualizing} so as to encourage the latent to follow a specific distribution (e.g. Gaussian).

\section{Discussion and Conclusion}
In this paper, we presented a new algorithm, AURORA, which combines a QD algorithm with a DR algorithm to enable robots to autonomously discover the range of their abilities. We have evaluated our approach on two experiments: a ballistic task and an air hockey scenario. In both scenarios, AURORA produced repertoires that cover a similar range of behaviors as those created with a certain amount of prior knowledge or domain expertise. Another useful aspect of AURORA is that it automatically defines the BD function, which is one of the most challenging aspects of QD algorithms~\cite{pugh2015confronting,cully2017quality,cully2018hierarchical}. 
Surprisingly, AURORA is not any slower (in terms of convergence speed) than approaches with fixed BD, even if it has to learn the BD function during the process. Overall, AURORA is a promising approach for scaling the QD algorithm to high-dimensional data, like CVT-MAP-Elites~\cite{vassiliades2018using}, but without the need for prior knowledge about the robot or the environment.

In future works, we will investigate how AURORA scales to more sophisticated robots (e.g., with many degrees of freedom or soft robots) and to more complex environments (e.g., with multiple objects). We will also investigate the use of multi-modal data, via the fusion for multiple sensory streams, and more complex DR algorithms, like recurrent LSTM-based auto-encoders~\cite{zhao2018robust}.  In particular, it will be interesting to study the curriculum generated by the robot when it has access to images of itself and its environment. We can imagine in the early stages of its exploration, AURORA will discover how to control the robot, and in the later stage, how to interact with the environment. 

While AURORA removes the need of domain knowledge to define the BD, several technical questions remain open. One of them regards the influence of the size of the latent representation. A larger latent space would probably increase the behavioral diversity, but might also affect the selective pressure in the QD algorithms. 
Finally, we will explore how AURORA can be extended to generate hierarchical behavioral repertoires~\cite{cully2018hierarchical}, such that AURORA will autonomously discover behaviors of increasing complexity.

\begin{spacing}{0.9}\small
\bibliographystyle{ACM-Reference-Format}
\bibliography{biblio} 
\end{spacing}

\end{document}